# Evolution of Autopoiesis and Multicellularity in the Game of Life

Peter D. Turney[*]

## Abstract

Recently we introduced a model of symbiosis, *Model-S*, based on the evolution of seed patterns in Conway's *Game of Life*. In the model, the fitness of a seed pattern is measured by one-on-one competitions in the *Immigration Game*, a two-player variation of the Game of Life. Our previous article showed that Model-S can serve as a highly abstract, simplified model of biological life: (1) The initial seed pattern is analogous to a genome. (2) The changes as the game runs are analogous to the development of the phenome. (3) Tournament selection in Model-S is analogous to natural selection in biology. (4) The Immigration Game in Model-S is analogous to competition in biology. (5) The first three layers in Model-S are analogous to biological reproduction. (6) The fusion of seed patterns in Model-S is analogous to symbiosis. The current article takes this analogy two steps further: (7) Autopoietic structures in the Game of Life (*still lifes*, *oscillators*, and *spaceships* — collectively known as *ashes*) are analogous to cells in biology. (8) The seed patterns in the Game of Life give rise to multiple, diverse, cooperating autopoietic structures, analogous to multicellular biological life. We use the *apgsearch* software (Ash Pattern Generator Search), developed by Adam Goucher for the study of ashes, to analyze autopoiesis and multicellularity in Model-S. We find that the fitness of evolved seed patterns in Model-S is highly correlated with the diversity and quantity of multicellular autopoietic structures.

**Keywords:** Evolution, autopoiesis, multicellularity, cellular automata, diversity, symbiosis.

---

[*] Ronin Institute, 127 Haddon Place, Montclair, NJ 07043-2314, USA,

peter.turney@ronininstitute.com, 819-661-4625.

# 1 Introduction

In previous work [26], we introduced *Model-S*, a computational model of the evolution of symbiosis. In Model-S, seed patterns compete for survival in one-on-one competitions in the *Immigration Game* (*IG*), invented by Don Woods in 1971 [12, 31]. The Immigration Game is a two-player variation on the solitaire *Game of Life* (*GoL*), invented by John Conway in 1970 [5, 22]. The winner of IG is the seed pattern that grows the most. Model-S adds an external evolutionary algorithm to IG. The algorithm evolves a population of seed patterns, selecting seeds according to their ability to win IG.

Model-S is constructed with four layers, each subsequent layer building on the previous layers. The purpose of having four layers is to measure each additional layer's contribution to the fitness of the population. This provides some insight into the strengths and weaknesses of each layer. Figure 1 presents a flow chart that sketches the basic structure of Model-S. (For more detail, see our previous article [26].)

Insert Figure 1 here.

Layer 1 implements a simple form of asexual reproduction, with a fixed genome size (that is, a fixed binary matrix size). A member of the population is selected for reproduction using tournament selection. The chosen seed pattern is mutated by randomly flipping some of the bits in the binary matrix and it then competes in a series of one-on-one Immigration Games with the other members of the population. Its fitness is the average fraction of games it wins. Layer 2 implements a slightly more sophisticated asexual reproduction. In addition to mutation by flipping bits (in Layer 1), the binary matrix is allowed to grow or shrink by adding or subtracting a row or column to or from the matrix. Layer 3 selects two seeds from the population and then combines them with genetic crossover. The resulting child is then passed on to Layer 2 for mutation. Layer 4 adds symbiosis to Model-S. Two seeds are selected from the population and fused together, side-by-side, creating a new symbiotic genome. This new fused seed is treated as a whole; that is, selection shifts from the level of the two parts to the level of the whole. The main result of our past work is evidence for the hypothesis that symbiosis promotes fitness improvements in IG [26].

In the course of our past research with Model-S, it became apparent that *autopoiesis* plays an important role in determining the fitness of seed patterns. The term *autopoiesis* was introduced by Humberto Maturana and Francisco Varela [19]. Randall Beer [2, p. 185] states that "… an autopoietic system consists of a network of processes that produce components whose interactions serve to generate and maintain the very network of processes that produced them."

Beer has shown, in a series of articles [1, 2, 3], that GoL is an excellent platform for modeling autopoiesis. The classes of structures in GoL that are most interesting to Beer, regarding autopoiesis, are *still lifes*, *oscillators*, and *spaceships* [3]. These three classes of patterns are structures that naturally emerge in the game and capture the attention of GoL players. A popular activity in the GoL community is to create a random *soup* and run it to quiescence [10]. A *soup* is a random initial seed pattern. When the soup reaches quiescence, what remains is called *ash*. Ashes are defined as still lifes, oscillators, and spaceships [11].

Figure 2 presents a sample of the output of Model-S. The left column of Figure 2 shows four initial seed patterns, randomly sampled from the final generations of four different runs of Model-S. The right column shows the final ash that results from each seed. The first row presents a run of Model-S with Layer 1 alone; the other layers are turned off. Each following row shows the effect of adding another layer to the model.

> Insert Figure 2 here.

Table 1 summarizes the types and quantities of ash for the four different runs of Model-S in Figure 2. Table 1 suggests that symbiosis (Layer 4) in Model-S yields large quantities of ash, compared with the other layers. In Section 4, we will present statistical analysis that supports the hypothesis that symbiosis yields highly fit seed patterns by evolving seeds that yield diverse types and large quantities of ash.

> Insert Table 1 here.

Model-S is a highly abstract, simplified model of biological life. Table 2 outlines a mapping between elements of biological life and elements of Model-S. Our previous article on Model-S [26] discussed the analogies presented here in the first six rows in Table 2: (1) The initial seed pattern in a cellular automaton is like a genome. It is a static encoding of the information that determines how the game will unfold, according to the rules of the game. (2) Running IG is like the growth of the phenome over time. Two seed patterns grow and compete with each other for space. (3) The tournament selection algorithm in Model-S is like natural selection. The winner of a tournament is selected for reproduction. (4) IG is used to measure the fitness of a player, based on the number of living cells in the player. IG is a model of the fitness of organisms competing for reproduction. (5) Layers 1, 2, and 3 in Model-S implement three different forms of reproduction: Layer 1 reproduces a selected genome with random binary mutations, Layer 2 reproduces a selected genome with changes to the size of the genome, and Layer 3 combines two genomes with genetic crossover. (6) The fusion of seed patterns in Layer 4 in Model-S is like symbiosis. In particular, fusion can be viewed as a form of endosymbiosis with vertical transmission of genes. The current article focuses on the last two rows in Table 2: (7) Ashes (still lifes, oscillators, and spaceships) are like living cells. (8) Multiple ashes that grow from a single seed pattern are like multiple cells in a living organism.

> Insert Table 2 here.

Before we continue, let's pause to discuss a possible source of confusion. The word *cell* might refer to a square in the grid of the Game of Life or it might refer to an autopoietic structure, such as ash in the Game of Life or an actual biological cell. When there might be confusion, we will use the terms *grid-cell* for a square in the grid and *ash-cell* for an autopoietic structure.

The main contributions of this article are (1) evidence that the fitness of a seed pattern depends on the number of ashes that it is able to generate (the *productivity* of the seed pattern, in Section 4.1), (2) evidence that the fitness of a seed pattern depends on the variety of ashes that it is able to generate (the *diversity* of types of ashes, in Section 4.2), and (3) evidence that, although the fitness of a seed pattern is correlated with its size, size alone is not sufficient for fitness: productivity and diversity require complex evolved structures (in Section 4.3).

These three lines of evidence support the analogies presented in the last two rows of Table 2. (1) A seed pattern that is highly productive is likely to win in competition against a seed pattern that is less productive. Likewise, in biology, an organism with many cells may have an advantage when competing against an organism with few cells. Model-S measures fitness using IG, where the winner is the pattern with the most living grid-cells, but productivity is based on the number of ash-cells. It does not necessarily follow that productivity (number of ash-cells) is correlated with fitness (number of grid-cells); therefore, an experiment is required to determine the relation between productivity and fitness (see Section 4.1). (2) A seed pattern that produces a diversity of types of ashes is likely to have an advantage over a seed pattern that has little variety, because each type of ash may require a different competitive strategy to defeat it. In biology, an organism with several types of cells (such as skin cells, muscle cells, and brain cells) may have an advantage when competing against an organism with only a few types of cells (see Section 4.2). (3) Winning IG requires complex evolved seed patterns, just as survival of biological organisms requires complex evolved genomes (see Section 4.3).

Although we are using Model-S as a specific example in this article, the phenomena we observe (increasing instances of autopoiesis and multicellularity) are a consequence of trying to improve the odds of winning the Immigration Game; the phenomena are not specific to Model-S. It seems necessary that any effort to create soups (seed patterns) that are increasingly better at winning Immigration Games will result in soups that yield increasing productivity and diversity of ashes. From this perspective, the current article tells us more about the nature of the Immigration Game and the Game of Life than it tells us about Model-S.

We review related work in Section 2. In Section 3, we present background information about Model-S (Section 3.1) and autopoiesis (Section 3.2). This background supports the core of the paper in Section 4, the experiments with fitness and productivity (Section 4.1), fitness and diversity (Section 4.2), the relations among productivity, diversity, and size (Section 4.3), and the lessons learned from the experiments (Section 4.4). Section 5 discusses future work and limitations of the analysis. The article concludes in Section 6.

## 2 Related Work

Conway's Game of Life was introduced to the world by Gardner in 1970 [5]. His introductory article mentioned *still lifes*, *oscillators*, and *spaceships*, the three classes of objects that are now covered by the term *ash* [11]. The phrase *primordial soup* was used by Wainwright in 1971 to describe a random initial state in GoL [6]. The GoL community generates random soups and runs them to quiescence, in order to find interesting ashes [9].

Varela and Maturana introduced *autopoiesis* to the world in 1973 [19]. McMullin [21] reports that Varela and Maturana had coined the word *autopoiesis* in 1971 and they were planning at that time to make a cellular automaton computer simulation of autopoiesis, inspired by the work of von Neumann [30] and Conway [5]. Their cellular automaton model was published in 1974 [21, 29]. It used a simulated chemistry in a two-dimensional grid space. A cell in the space could be empty or it could contain one of three types of particles. The rules of the artificial chemistry were designed to allow particles to connect and form a kind of cell membrane. The membrane was capable of self-maintenance and self-construction.

Beer has made a strong case for GoL as a model of autopoiesis [1, 2, 3]. At first, he focused on *gliders* as instances of autopoiesis [1, 2], but later he expanded his studies to *still lifes*, *oscillators*, and *spaceships* [3]. This expansion connects the research in autopoiesis to the investigations of the GoL community, exploring soups and ashes.

The term *symbiosis* is derived from the Ancient Greek word συμβίωσις (sumbíōsis), meaning *living together*. There are many different types of symbiosis [4]. Martin and Schwab write [18, p. 32], "Confusion has afflicted the definition of *symbiosis* for over 130 years." They propose a set of terms based on whether the two species in a symbiotic relationship experience a beneficial effect (+), a harmful effect (-), or a neutral effect (0). This yields six types of symbiosis: *neutralism* (0/0), *antagonism* (-/-), *amensalism* (0/-), *agonism* (+/-), *commensalism* (+/0), and *mutualism* (+/+) [18, see their Figure 10]. They also distinguish *endosymbiosis* (one organism lives within the tissues of the other) and *ectosymbiosis* (one organism lives on the body surface of the other, including the digestive tract).

Maynard Smith and Szathmáry, studying the major transitions in evolution, found that several of the major transitions involved symbiosis with a shift in the level of selection [20]. The prototypical example is the transition from prokaryotes to eukaryotes [16, 17, 20]. This is an instance of endosymbiosis where the two organisms became so tightly bound together that they cannot live apart, so natural selection must select them as a whole; it is no longer possible to select only one of the two parts. Layer 4 in Model-S (fusion of seed patterns) is a model of endosymbiosis.

The current article brings together the work of the Game of Life community on soups and ashes [8, 9, 10], the research in theoretical biology on autopoiesis [2, 3, 29], the research of the evolutionary computation community on models of evolution [23, 32, 33], and the research in theoretical biology on symbiosis [4, 16, 17].

# 3 Background: Model-S and Autopoiesis

Before we present the experiments (in Section 4), we need to provide some more background information on Model-S (Section 3.1) and autopoiesis (Section 3.2).

## 3.1 Model-S

The Game of Life is a solitaire game in which the player creates an initial seed pattern in an infinite grid of square cells [5, 22]. The rules of GoL determine how the pattern changes over time. In Model-S, we view the initial seed pattern as analogous to a *genome* and the rules determine how the *phenome* develops from the genome (*ontogeny* — the development of an organism over its lifespan).

Each cell in the GoL grid is either *dead* (state 0) or *alive* (state 1). The state of a cell depends on the state of its eight nearest neighbours in the grid. The rules for GoL are concisely expressed as B3/S23: A cell is *born* (it switches from state 0 to state 1) if it has exactly three live neighbours (B3). A cell *survives* (it stays in state 1) if it has either two or three live neighbours (S23). Otherwise the cell remains dead or it dies (state 0).

Time passes in a series of discrete intervals. The states of the cells in the grid at time $t = 0$ constitute the initial seed pattern for the game. The states at time $t$ uniquely determine the new states at time $t + 1$. With each time step, all of the cells in the grid are updated.

The Immigration Game is a two-player game in which the two players each create an initial seed pattern [12, 31]. The rules for life and death are B3/S23, as in GoL, except that there are two *alive* states in IG (states 1 and 2). Conventionally, the background (state 0) is usually white and the living states (states 1 and 2) are usually red and blue. There are two new rules for determining colour: (1) Live cells do not change colour unless they die (switch to state 0). (2) When a new cell is born, it takes the colour of the majority of its live neighbours. Since birth requires three live neighbours, there is always a clear majority.

Model-S [28] uses the Golly cellular automata software [25] to run Immigration Games [12, 31]. In Model-S, Golly is controlled by Python scripts that implement evolution outside of Golly. A seed pattern is represented in Python by a binary matrix. The initial population consists of 200 random seed patterns. The algorithm for evolution is a steady-state model with a constant population size [32, 33], in which each new child replaces the least fit member of the population. Members of the population are chosen for reproduction using tournament selection [33].

In Model-S, fitness is determined by one-on-one competitions between two seed patterns in IG. The seeds are represented as binary matrices when they are not competing. When it is time to compete, one seed is assigned red (state 1 remains state 1) and the other seed is assigned blue (state 1 switches to state 2). The two competitors are randomly situated in a toroidal grid with some space separating them. The game runs until a time limit is reached. The winner is the seed that grows the most. The growth of the red seed is the number of cells in state 1 at the end of the game minus the number of cells in state 1 at the beginning of the game. Likewise, the growth of the blue seed is the increase in blue states. If a seed shrinks instead of growing, its score is zero. Ties are allowed.

In these competitions, the size of the toroidal grid and the time limit for the game are both functions of the sizes of the seeds. Larger seeds are allowed larger grids and longer time limits. The reasoning is that any fixed limit on space and time would set a bound on the possible evolution of the seeds over many generations, which would be less interesting than unbounded growth. The motivation for a toroidal grid is to limit the space for growth, so that the two seeds are forced to interact. Without a limit on space, the two seeds might simply avoid each other, which would be less interesting than interaction. The toroid limits the space for any given competition, but the toroid increases in size from one competition to another, if the seed patterns increase in size. Likewise, the time limit is fixed for any given competition, but can increase from one competition to another. (Space and time can also decrease, if the seeds happen to decrease in size.)

In IG, if states 1 and 2 were coloured black, the game would appear to be exactly the same as GoL. The purpose of the two colours is to score the two players, to convert GoL from a solitaire game into a two-player competitive game. In IG, we can choose to be red-blue colour blind, and view the game as GoL. In the work we describe in this article, Model-S measures the fitness of seeds by competitions in the three-state Immigration Game (two seeds at a time), but the ashes are analyzed in the two-state Game of Life (one seed at a time).

The fitness of a seed in IG is the fraction of games that it wins, playing against every seed in the population. The initial population consists of 200 randomly generated seed patterns. Tournament selection randomly chooses two seeds from the population and the fitter of the two is selected for reproduction. The type of reproduction depends on which layers are enabled. The new child replaces the least fit member of the population, so the population remains at 200 for the entire run of Model-S. When 200 children have been born, we say that one generation has passed. We run Model-S for 100 generations, resulting in 20,000 births (100 generations with 200 births = 20,000 births; the generation count goes from 0 to 100, but generation 0 is randomly generated, not born from parents).

The fitness of a seed pattern is measured by two competitions against every seed pattern in the population (2 competitions per competitor × 200 competitors = 400 competitions, including 2 self-competitions). Each competition involves two different seed patterns in the same grid, playing IG. The two seed patterns interact with each other, which means that their phenomes are different in each competition, due to their interaction. The *fitness* of a seed pattern is defined as the average number of games the pattern wins, playing IG. This average is not a property of the phenomes, since each phenome is different. Fitness is a property of the given genome (the seed pattern that is currently being evaluated), which is the same in each competition. On the other hand, the *productivity* of a seed pattern is measured with a single seed pattern in a grid, playing GoL. There is no need to repeat runs, since the final ashes will always be the same. We could say that productivity and diversity are a product of the seed pattern (genome) or a property of the ashes (phenome).

The fitness in Model-S is an internal, relative fitness. The fitness of a seed is relative to the fitness of the other seeds in the current population. With each new birth, the fitness of every seed in the population is updated. It follows that the average fitness (the average fraction of games won) is always 0.5. This means that we cannot compare the fitness of two seeds that are sampled from different populations. We argued in our prior work [26] that an internal, relative fitness measure is more likely to result in open-ended evolution [24] than an external, absolute fitness measure, although we have not yet tested this hypothesis.

In order to evaluate the results from running Model-S, we need to create an external, absolute fitness measure that will allow us to compare seeds from different populations. This external fitness measure has no impact on the course of evolution in Model-S; it is only applied after a run of Model-S, in order to allow us to interpret the results of the run. The external, absolute measure we use is the fraction of Immigration Games that a seed wins when competing against seeds of the same size (the same number of rows and columns) and the same density (the same number of living cells in the initial seed pattern) [26]. A given seed's competitors are generated by copying the given seed and then randomly shuffling the cells in the copied seed. This ensures that the winner is winning due to the structure of the seed (the specific pattern of living and dead cells), not the size or density of the seed. This is analogous to the sport of boxing, where fighters are matched by weight.

Figure 3 shows the external, absolute fitness of the elite seeds in the final generation for each of the four layers of Model-S (blue bars). It also shows the shows the areas (rows × columns) of the elite seeds in the final generation for the four layers of Model-S (red bars). Each bar is the average of 12 separate runs of Model-S. For each generation (each 200 births), we sampled the top 50 most fit seeds (based on their internal, relative fitness) and calculate their external, absolute fitness. We call these seeds the *elite* seeds of the given generation. We saved all of the elite seeds from our past work, which allows us to use these seeds again in the current article.

Insert Figure 3 here.

We can see in Figure 3 that area is highly correlated with fitness (the correlation is 0.811 and it is statistically significant), but increasing area does not cause increasing fitness. Increasing area is necessary for increasing fitness, because increasingly complex structures require larger seeds to store increasing amounts of information. But increasing area is not sufficient for increasing fitness, because a specific kind of structure is required for fitness; that is, a structure that grows quickly and robustly during IG.

In our previous paper [26], introducing Model-S, we presented three different arguments in support of the claim that increasing area does not cause increasing fitness. We will only mention one of those arguments here: Note that the external, absolute fitness is defined by competitions between seeds that are matched in size and density. Therefore area cannot explain why the average symbiotic seed in the final generation of Layer 4 has a fitness of 93.6% (see Figure 3). If area were the deciding factor, the fitness would have to be 50%, because the two competitors have the same area. The only difference between the symbiotic seed and its competitor is that the competitor has been shuffled, which damages its structure but preserves its area. The structure is what determines the winner, not area.

## 3.2 Autopoiesis

There have been several automated censuses of ashes, counting the number of times each ash type appears in soups [10]. In the context of the Game of Life, it seems that the term *primordial soup* was first used by Robert Wainwright in 1971 [6]. More recently, Andrzej Okrasinski made a census of the ashes arising from random soups from 2003 to 2008 [13], Achim Flammenkamp made a census in 2004 [14], Nathaniel

Johnston made a census from 2009 to 2011 [15], and Adam Goucher has been collecting census information since 2015, using his *apgsearch* software (Ash Pattern Generator Search) [8].

Goucher's apgsearch is currently the most popular tool for running random soups and classifying their ashes. Apgsearch was designed for distributed computation, with many different users on the Internet running random soups and uploading their ash censuses to a central server. Over 100 users have contributed to the census of ash objects [9].

Our previous work with Model-S [26] demonstrated that Layer 4, the symbiotic layer, when combined with the other three layers, resulted in a significant increase in the fitness of the population. In that prior work, we did not analyze the specific means by which symbiosis achieved greater fitness, except to provide evidence that the increased fitness was not merely due to increased size in the initial seed pattern. We conjectured that autopoiesis was somehow responsible [26], but we did not know how to test this hypothesis at that time. The solution became clear when we discovered Goucher's apgsearch software [8].

In the following experiments, we use a slightly modified version of Goucher's apgsearch to identify and count the ash that is generated by evolved seed patterns from Model-S [28]. The modification allows us to gather a census of ash for a single, specific seed pattern, instead of accumulating counts from a large number of random soups, as the original version of apgsearch does.

Apgsearch is currently at version 5.0, which is written in C++. We used apgsearch version 1.1, which is written in Python and uses Golly, because Model-S is also written in Python and uses Golly. We modified this version of apgsearch to analyze the ash from a single seed pattern, rather than aggregating ash statistics from very large samples of soups. The Python version of apgsearch is slower than the C++ version, but speed was not an issue for our analysis of Model-S seeds.

Apgsearch has discovered 159,347 distinct types of ash objects so far [9]. Beer has studied three of these objects (a block, a blinker, and a glider) in depth, to determine whether they satisfy Maturana and Varela's two conditions for autopoiesis, the *closure condition* and the *boundary condition* [3, p. 2]:

> "The *closure condition* demands that the network of processes must produce the components whose interactions generate and maintain that very same network. The *boundary condition* demands that the spatial boundary that distinguishes an autopoietic system from its background must itself be generated and maintained by the network of processes and in turn must play a central role in enabling those same processes."

Beer finds that all three objects (blocks, blinkers, and gliders) satisfy the two conditions for autopoiesis [3].

Blocks, blinkers, and gliders are the simplest members of their respective classes, still lifes, oscillators, and spaceships. Beer does not explicitly state that all still lifes, oscillators, and spaceships satisfy the conditions for autopoiesis, but he hints that he believes this is true [3, p. 15]: "This article has argued that *such GoL entities as* [emphasis added] blocks, blinkers, and gliders are autopoietic with respect to the GoL physics because they satisfy both the closure and boundary conditions of Maturana and Varela's definition." We believe that Beer's analysis generalizes to all still lifes, oscillators, and spaceships (that is, all ashes), but we do not have a formal proof of this. We will proceed with the working hypothesis that Beer's analysis generalizes to all ashes, in the expectation that the core ideas of our article will survive even if the working hypothesis may require some fine-tuning.

## 4 Experiments: Fitness, Productivity, Diversity, and Area

In this section, we present three new analyses of the data from our previous work with Model-S [26]. In Section 4.1, we apply the modified apgsearch to the stored elite seeds from our past work, in order to count the number of autopoietic structures that are produced (the *quantity* of ashes). In Section 4.2, we count the number of types of autopoietic structures the seeds produce (the *diversity* of ashes). In Section 4.3, we *shuffle* the seeds before we run apgsearch, to see whether the quantity and diversity depend on the size of the seeds or the structure of the seeds (shuffling changes structure but has no effect on size). The results show that the quantity and diversity of ash for a given layer of Model-S is highly correlated with the fitness of that layer. Shuffling greatly reduces the quantity and diversity of ashes, indicating that the quantity and diversity of ash is strongly affected by the structure of the seeds. Section 4.4 summarizes the results.

### 4.1 Fitness and Productivity

The *productivity* of a seed pattern is the quantity of ash it generates. For example, the initial seed pattern for Layer 4 in Figure 2 yields 106 objects, as we can see in Table 1 (bottom right corner). For each of the four layers of Model-S, Figure 4 shows the average quantity of ash produced by seed patterns in the final generation. Each red bar in Figure 4 is the average productivity of 600 seed patterns (50 elite seeds $\times$ 12 runs of Model-S).

Insert Figure 4 here.

If we compare the productivity (red bars) of the four layers in Figure 4 with the fitness (blue bars), we see the same general pattern in productivity and fitness: The productivity and fitness of Layer 1 (the uniform asexual layer) is low. Layer 2 (the variable asexual layer) and Layer 3 (the sexual layer) have similar levels of productivity and fitness, with a slight advantage to Layer 2. Both Layers 2 and 3 are more productive and fit than Layer 1. Layer 4 (the symbiotic layer) has the highest productivity and fitness. Figure 4 shows that the symbiotic layer is much more productive than the other layers and it is much more fit. The correlation between fitness and productivity in the final generation is high (0.706) and statistically significant.

The average seed pattern in the final generation of Layer 4 (the symbiotic layer) generates 73.1 ashes (see Figure 4) and has an area of 94.1 (see Figure 3). Therefore the average ash generated by the final generation of symbiotic seeds is 0.78 ashes per unit area (73.1 ashes / 94.1 area).

Catagolue reported a census on April 27, 2019 of 18,928,504,510,982 random soups, resulting in a total of 413,493,174,300,923 ashes [9]. This yields an average of 21.8 ashes per random soup. Each Catagolue soup begins with a random 16×16 binary matrix, which has an area of 256 cells. Therefore the average ash generated by random soups is 0.09 ashes per unit area (21.8 ashes / 256 area).

The average final generation seed pattern of Layer 4 has an area that is 37% (94.1 area / 256 area) of the area of Catagolue's random soups and a productivity that is 335% (73.1 ashes / 21.8 ashes) of the productivity of Catagolue's soups. Therefore the evolved seeds of Layer 4 are 9.1 times more productive than random soups (335% / 37%). This shows that the evolved seeds in the final generation of Layer 4 are much more productive than random seeds. However, Model-S was not designed to produce ash; it was designed to play IG. With hindsight, it makes sense that winning IG requires evolution to generate seeds that maximize the production of autopoietic structures.

## 4.2 Fitness and Diversity

The *diversity* of a seed pattern is the number of types of ashes it generates. For example, the initial seed pattern for Layer 4 in Figure 2 yields 10 types of ashes, as we can see in Table 1 (bottom right corner). Figure 5 shows the average diversity of ashes produced by seed patterns in the final generation of Model-S, for each of the four layers. Each red bar in Figure 5 is an average of 600 seed patterns (50 elite seeds × 12 runs of Model-S).

Insert Figure 5 here.

If we compare the diversity (red bars) of the four layers in Figure 5 with the fitness (blue bars), we see the same general pattern in diversity and fitness. This qualitative similarity between diversity and fitness is confirmed by measuring the correlations. The correlation between diversity and fitness is 0.834. The correlation is high and it is statistically significant.

We know from Section 4.1 that the evolved seed patterns in Layer 4 are much more productive than random soups (there is a large quantitative difference between the number of seeds generated in Layer 4 and the number of seeds generated in random soups — a factor of 9.1), but is there a difference in the types of seeds in Layer 4 and the types of seeds in random soups (is there a qualitative difference in the distribution of seed types)?

Table 3 shows all of the types of ash found in each layer of Model-S in the final generation. Focusing on Layer 1 as an example (the first column in Table 3), there are seven types of ash, ranked in descending order of their frequencies in their layer in Model-S; that is, the *block* is the most frequent ash found in Layer 1 and the *boat* is the least frequent ash found in Layer 1. For each ash type in the first column, there is a corresponding number in the second column that gives the rank of the ash type in Catagolue [7] (from a census in July 15, 2017). For example, the *block* is ranked 1 in Catagolue and the *block* is also first in Layer 1. However, the *pond* is third in Layer 1, yet it is ranked 9 in Catagolue.

Insert Table 3 here.

For most entries in Table 3, the rank of an ash type that occurs in Model-S is approximately the same as its rank in Catagolue. The *boat* is the seventh ash type in Layer 1 (see column 1) and it is the sixth ash type in Catagolue (see column 2). The *tub* is the eighth ash type in Layer 4 (see column 7) and it is the eighth type in Catagolue (see column 8). However, although the ranks of ash types in Model-S are mostly similar to their ranks in Catagolue, there are a few outliers. In Table 3, we have highlighted in italics one outlier in each of the four layers: the *pond* in Layers 1 and 3 (ranked third in Model-S but ranked ninth in Catagolue), the *pulsar* in Layer 2 (ranked tenth in Model-S but ranked twenty-first in Catagolue), and the *loop* in Layer 4 (ranked twelfth in Model-S but ranked forty-ninth in Catagolue). We discuss this topic further in Section 5.

## 4.3 Productivity, Diversity, and Area

In Section 3.1, we argued that, although the fitness of a seed pattern (its tendency to win Immigration Games) is correlated with the area of the seed pattern (see Figure 3), area does not cause fitness; area merely provides room for complex structure (area allows fitness), and fitness is the result of a certain type of complex structure (the type that wins games). In the current section, we present two experiments that further support the claim that increasing area is not sufficient for increasing fitness. Figure 6 compares the productivity of the seeds before shuffling (the *intact* bar) and after shuffling (the *shuffled* bar). The productivity for each bar in Figure 6 is the average of 600 values (12 runs × 50 elite seeds). The productivity of the shuffled seeds ranges from 7% to 14% of the productivity of the intact seeds. This is evidence that the kind of structure that yields fitness is also the kind of structure that yields productivity. By design, shuffling has no impact on area and density, so the loss in productivity that we see here has nothing to do with area or density; it can only be caused by the change in structure. The shuffled seeds in Layer 4 are less productive than the intact seeds in Layer 1 (7.18 for Layer 4 shuffled versus 13.00 for Layer 1 intact — see Figure 6), although the seeds in Layer 4 have much greater area than the seeds in Layer 1 (see Figure 3).

Insert Figure 6 here.

Figure 7 compares the diversity of the seeds before shuffling (*intact*) and after shuffling (*shuffled*). The diversity of the shuffled seeds ranges from 18% to 33% of the diversity of intact seeds. Shuffling clearly reduces diversity, although the reduction is likely to be a consequence of the reduced productivity of the shuffled seeds. As the number of ashes decreases, the number of types of ashes also tends to decrease, so the reduced diversity is likely a side effect of reduced productivity. The shuffled seeds in Layer 4 are less diverse than the intact seeds in Layer 1 (2.57 for Layer 4 shuffled versus 3.05 for Layer 1 intact — see Figure 7), although the seeds in Layer 4 have much greater area than the seeds in Layer 1 (see Figure 3).

Insert Figure 7 here.

The results in this section support the hypothesis that area alone is not sufficient for a seed pattern to win IG. Area is required to encode complex seed patterns, but area is not sufficient for high fitness. In the next section, we consider what is necessary for fitness.

## 4.4 Summary of Results

Table 4 shows the correlation between fitness and three other attributes: area, productivity, and diversity. The table provides evidence that fitness is more correlated with diversity than it is with area or productivity.

Insert Table 4 here.

The results in Section 4.3 (see Figure 6 and Figure 7) tell us that, although increasing area is *necessary* for increasing fitness, increasing area *does not cause* increasing fitness. Increasing area combined with increasing diversity appears to be the mechanism by which evolution in Model-S achieves high fitness in Layer 4. A wide range of types of ashes yields a high level of fitness, and a wide range of types of ashes requires increasing area to accommodate new types.

In our previous work with Model-S [26], we demonstrated that Layer 4 (symbiosis) was able to evolve seed patterns with high fitness, but we did not explain what made these seed patterns more fit than other patterns. A contribution of the current article is the insight that fitness is the result of a diversity of autopoietic structures. This supports our view that Model-S is analogous to biological life in eight ways (see Table 2), whereas our previous work only supported the first six of these eight correspondences between characteristics of biological life and Model-S.

## 5 Future Work and Limitations

In Section 4.2, we noted that *ponds*, *pulsars*, and *loops* seem to be more frequent in Model-S than we would expect from Catagolue. It seems possible that they are selected because they are more stable than other objects, so they can survive close contact with other objects while playing IG, whereas other objects are destroyed. We plan to test the stability of these objects, relative to other objects, and to look for other properties that might explain why *ponds*, *pulsars*, and *loops* seem to be favoured.

The score in IG is based on growth, which is measured by the increase in the number of living cells of a given colour (red or blue). In Section 4.1, we counted the number of ash-cells (autopoietic structures), which is not the same as counting the number of living grid-cells (squares in the grid). Selection in Model-S might be biased towards producing either many small ash-cells or a few large ash-cells: For example, many small ash-cells might be less vulnerable in IG than a few large ash-cells. We plan to look for a bias towards either smaller ash-cells or larger ash-cells. With random soups, we see there is a bias towards smaller ash-cells; smaller ash-cells seem to occur more frequently than larger ash-cells [7]. To see whether IG prefers smaller ash-cells, we need to take into account this natural bias for smaller ash-cells.

One possibility would be to alter IG so that the winner is determined by the number of ash-cells, instead of the number of living grid-cells. If this alteration changes the distribution of ash types, then we will learn something about how the method of scoring affects the kinds of ash we see.

Another area for future work is to explore other cellular automata rules, besides GoL, to see how well the behaviours we have observed for GoL generalize to other rules. The rules for GoL belong to the family of semitotalistic rules, a family with 262,144 members. Woods' method for converting GoL into a two-player game generalizes to 8,192 members of the family of semitotalistic rules. We estimate that about 4,000 of these rules will show the kind open-ended evolution that we have seen with Woods' Immigration Game [27].

We mentioned in Section 3.2 that it has not yet been proven that all still lifes, oscillators, and spaceships satisfy the conditions for autopoiesis. It seems likely that this is true, but the proof might be complicated. We leave this for future work.

# 6 Conclusion

Our previous article on Model-S [26] presented a highly abstract and simplified model of six aspects of biological life: (1) genomes, (2) phenomes, (3) natural selection, (4) competition, (5) reproduction, and (6) mutualism. These shared abstract elements are given in the current article in first six rows in Table 2. The previous article mentioned autopoiesis, but only as a topic for future work. The current article presents that promised future work, adding (7) cells and (8) multicellular organisms — the last two rows in Table 2. The key that made this extension possible was Goucher's apgsearch software [8], which allows us to perform a census of ash in experiments with Model-S. The apgsearch software permitted precise measurements of the productivity and diversity of large quantities of ash (Section 4).

The experiments in Section 4 go beyond our previous work [26] by providing some insight into how Layer 4 is able to attain high levels of fitness. Past work demonstrated that certain specific and relatively rare initial seed patterns were able to consistently win Immigration Games, but we could not say exactly what property of these seed patterns enabled their success. Now we can say that successful seed patterns are those that create a diversity of autopoietic structures (Figure 5 in Section 4.2). It is likely that there is much more that can be said about successful seed patterns, but we leave that as future work. Our hope is that more research will increase our understanding of the analogies between evolving cellular automata and evolving biological life.

## Acknowledgments

Thanks to the *Artificial Life* reviewers for their very helpful comments on the article. Thanks to Andrew Trevorrow, Tom Rokicki, Tim Hutton, Dave Greene, Jason Summers, Maks Verver, Robert Munafo, Brenton Bostick, and Chris Rowett, for developing *Golly*. Thanks to Adam Goucher for developing *apgsearch*.

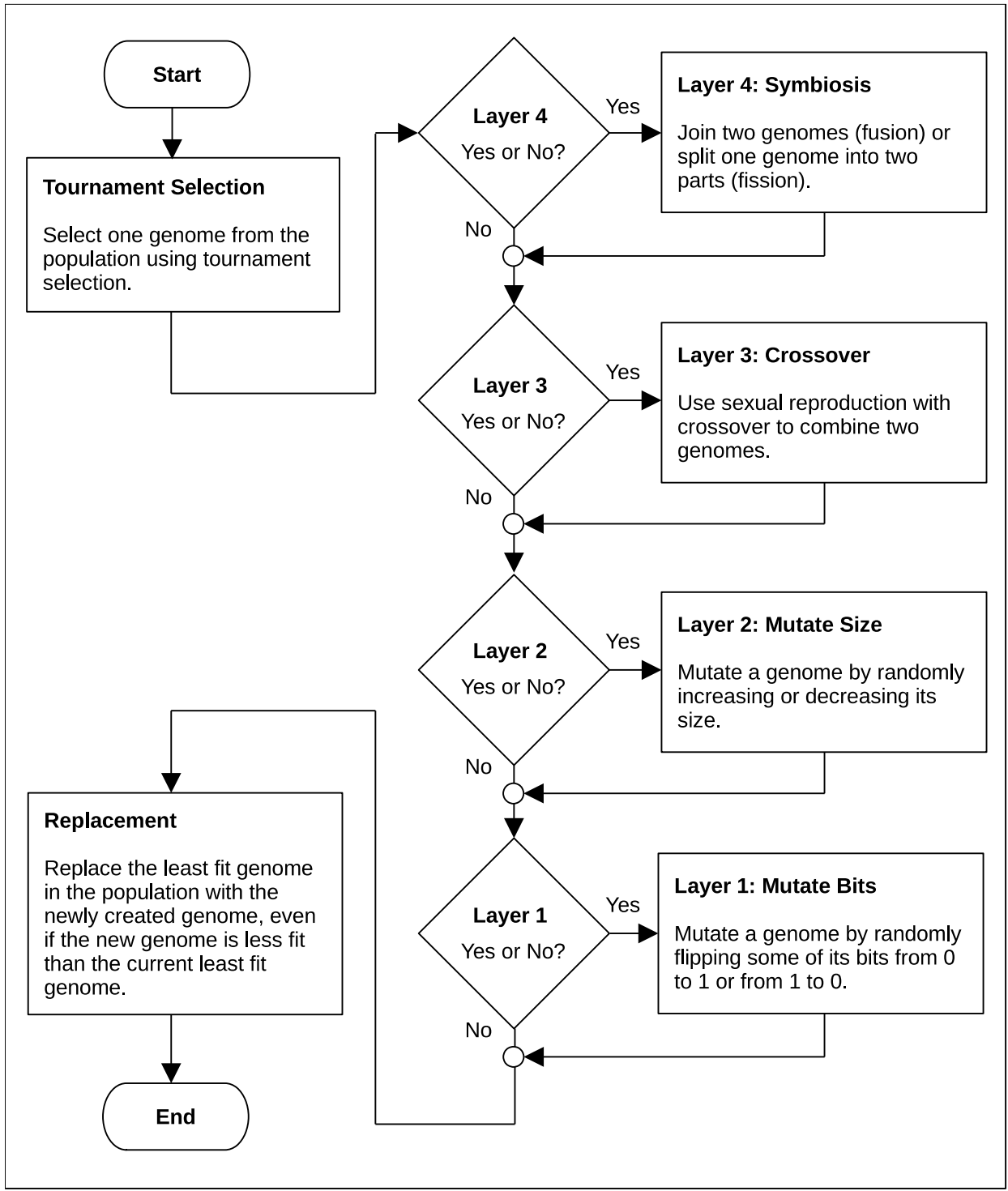

Figure 1. This flowchart outlines the process for selecting an individual's genome from the population and creating a new genome. This process is a subroutine in a loop that produces a series of new individuals. For each individual that is added to the population, another is removed; hence this is a steady-state model with a constant population size. The decision to use a given layer is determined by the parameters of Model-S.

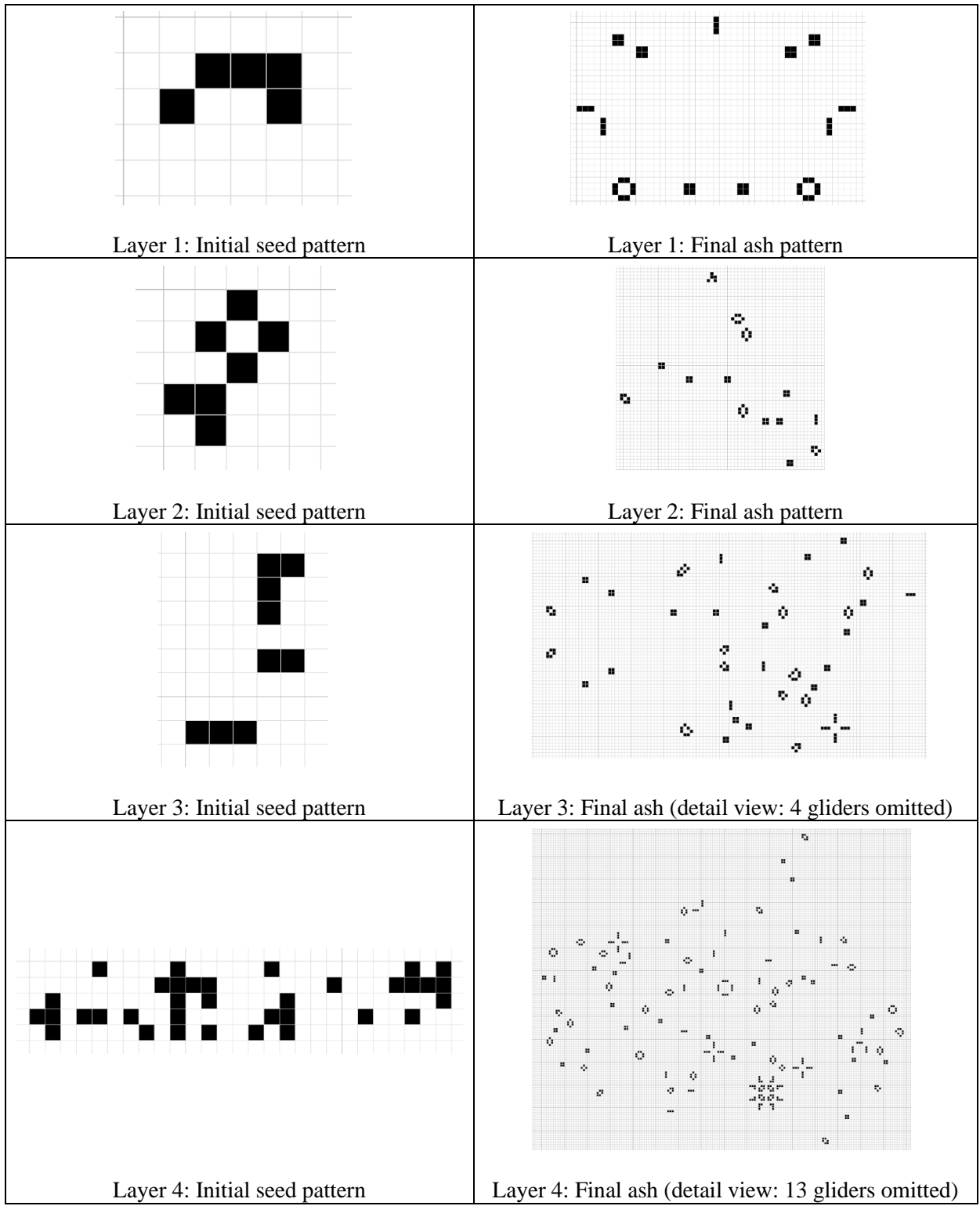

Figure 2. These images show examples of initial evolved seed patterns for each layer on the left and the corresponding final patterns on the right. The final patterns continue to change, but they are composed entirely of ashes. Gliders are omitted in Layers 3 and 4 because they have moved far from the core ashes.

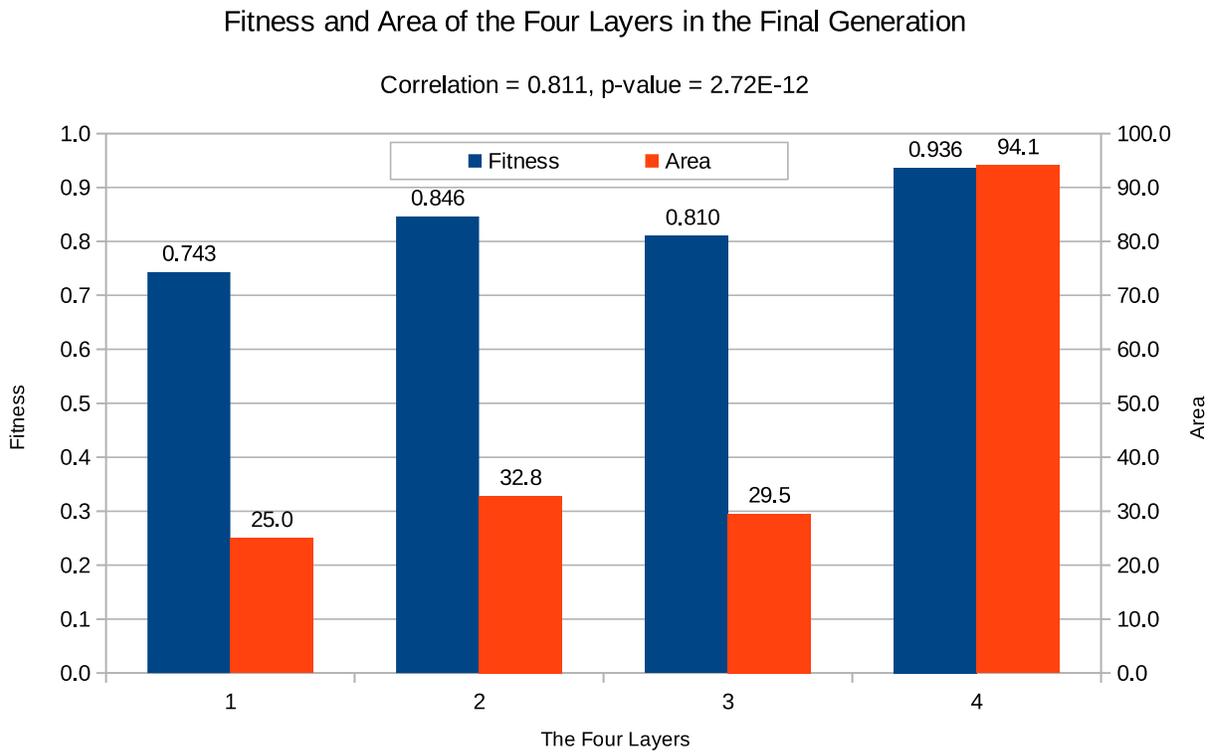

Figure 3. This bar graph presents the correlation between elite seed fitness in the final generation and elite seed area. The area of an elite seed is measured by the product of the width and height of the seed. The correlation between fitness and area (0.811) is high and significant. We evaluate the statistical significance of the correlation using a two-tailed Student t-test for Pearson correlations. The correlation is based on comparing two samples of 48 values each (4 layers × 12 runs).

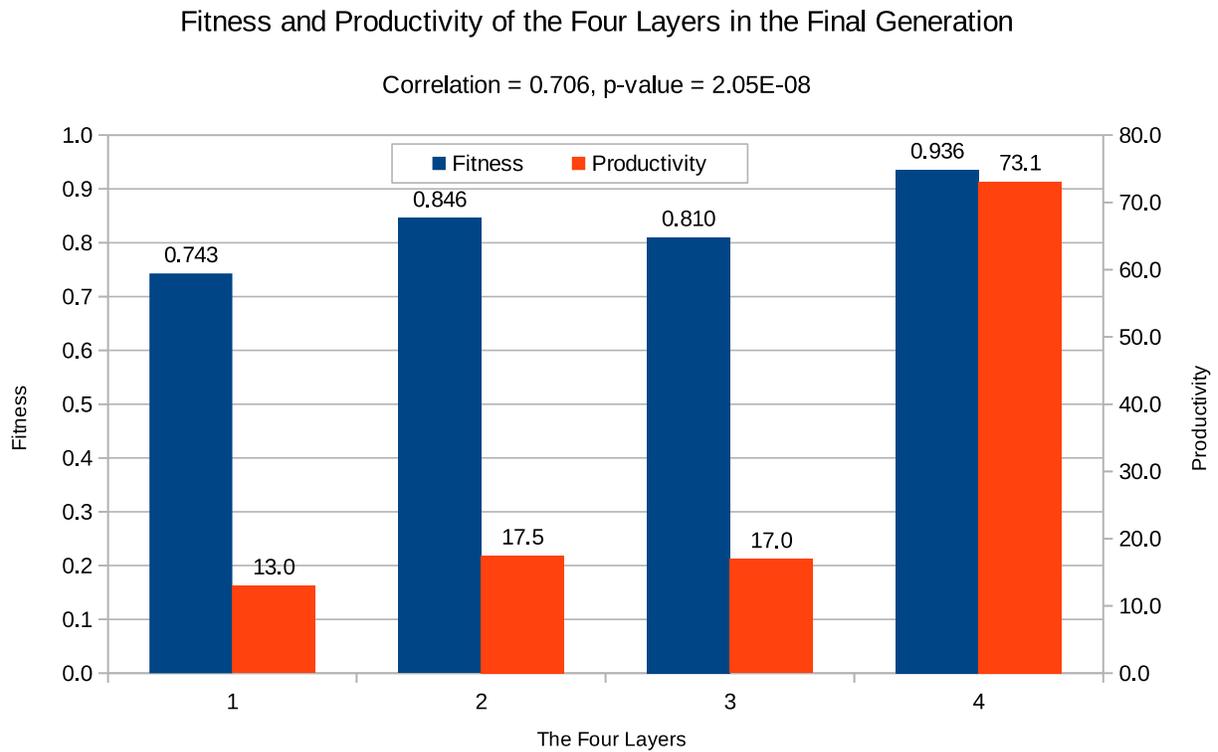

Figure 4. This bar graph presents the correlation between elite seed fitness in the final generation and elite seed productivity in the final generation. The productivity of an elite seed is measured by the number of ashes it creates. The correlation between fitness and productivity (0.706) is high and significant. We evaluate the statistical significance of the correlation using a two-tailed Student t-test for Pearson correlations. The correlation is based on comparing two samples of 48 values each (4 layers × 12 runs).

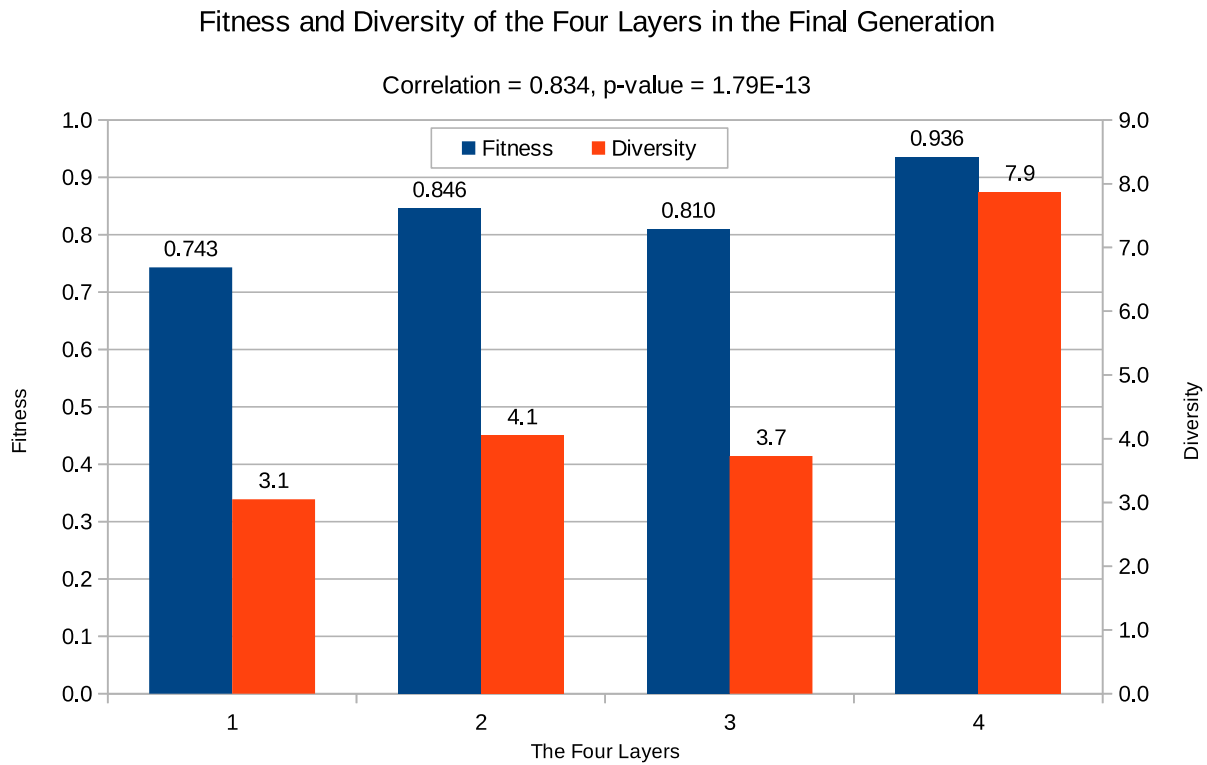

Figure 5. This bar graph presents the correlation between elite seed fitness in the final generation and elite seed diversity in the final generation. The fitness of an elite seed is measured by competitions between the elite seed and random seeds of matched size and density. The diversity of an elite seed is measured by the number of different types of ash that it yields. The correlation between fitness and diversity (0.834) is high and significant. We evaluate the statistical significance of the correlation using a two-tailed Student t-test for Pearson correlations. The correlation is based on comparing two samples of 48 values each (4 layers × 12 runs).

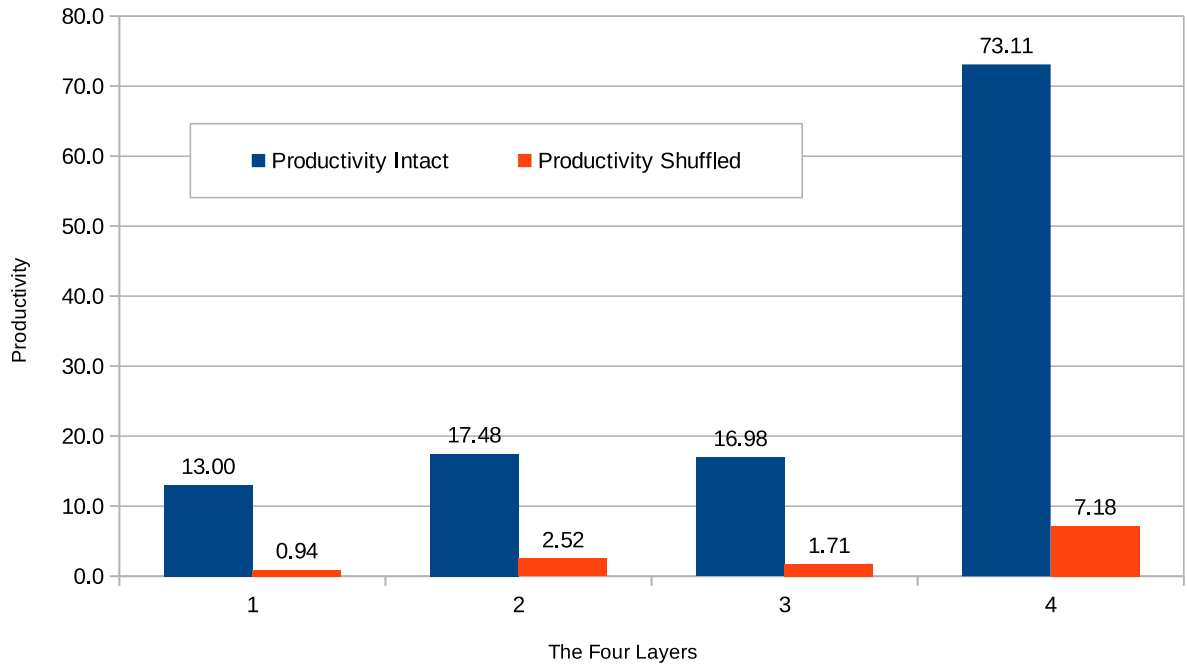

Figure 6. Here we compare the productivity of the final generations of the four layers with and without shuffling. The bar graph shows that shuffling greatly reduces productivity. The key factor for productivity is the specific evolved structure of the seed pattern, not the area of the pattern.

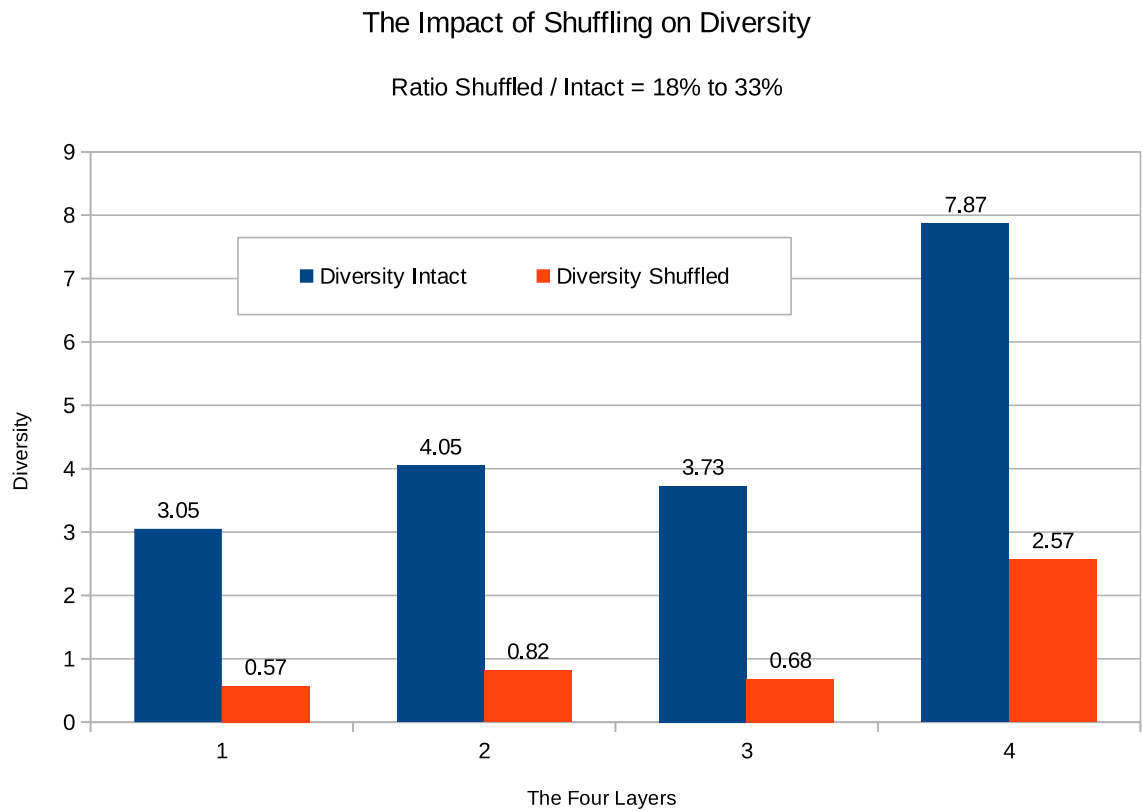

Figure 7. Here we compare the diversity of the final generations of the four layers with and without shuffling. The bar graph shows that shuffling greatly reduces diversity. This supports the claim that, although increasing area is required for increasing diversity, the key factor for diversity is the specific evolved structure of the seed pattern.

Table 1. This table summarizes the statistics for the final ash patterns shown in Figure 2. For more information about the types of ashes listed in the table, see Adam Goucher's Catagolue [7].

| Layer 1 ashes | | Layer 2 ashes | | Layer 3 ashes | | Layer 4 ashes | |
|---|---|---|---|---|---|---|---|
| Type | Count | Type | Count | Type | Count | Type | Count |
| block | 6 | block | 7 | block | 16 | blinker | 36 |
| blinker | 5 | beehive | 3 | blinker | 8 | block | 25 |
| pond | 2 | blinker | 1 | boat | 5 | beehive | 15 |
| | | glider | 1 | glider | 4 | glider | 13 |
| | | ship | 1 | beehive | 4 | boat | 6 |
| | | boat | 1 | ship | 2 | ship | 4 |
| | | | | loaf | 2 | pond | 3 |
| | | | | long boat | 1 | tub | 2 |
| | | | | | | loaf | 1 |
| | | | | | | pulsar | 1 |
| Num objects | 13 | Num objects | 14 | Num objects | 42 | Num objects | 106 |
| Num types | 3 | Num types | 6 | Num types | 8 | Num types | 10 |

Table 2. This table shows correspondences between characteristics of biological life and Model-S. In a highly abstract and simplified way, Model-S exhibits several of the main concepts of biology.

|   | Biological Life | Model-S | Shared Characteristics |
|---|---|---|---|
| 1 | genome, DNA, chromosomes | initial seed pattern, initial soup | static information, code |
| 2 | phenome, organism, development | playing the game, running the cellular automaton | dynamic development, growth |
| 3 | natural selection | tournament selection | selection |
| 4 | competition | the Immigration Game | fitness measure |
| 5 | reproduction with mutation, variation, sex | Layers 1, 2, and 3 of Model-S | reproduction with heritable variation |
| 6 | mutualism, cooperation | Layer 4, fusion of seed patterns | symbiosis |
| 7 | cells | still lifes, oscillators, spaceships, ashes | autopoiesis |
| 8 | multicellular organisms | multiple ashes grown from the same initial seed pattern | multicellularity |

Table 3. This table lists the types of ash that are created in each layer of Model-S. For each layer, we collect all the ashes generated from the 600 seed patterns in generation 100 (50 elite seeds × 12 runs). For each type of ash in each layer of Model-S, we rank the ashes in order of decreasing frequency and we show the rank of the type in Catagolue [7], where ashes are also ranked in order of decreasing frequency. In general, the rank of each ash type in Model-S is similar to the rank in Catagolue, but we have marked some possible outliers in italics.

| Layer 1 | Rank | Layer 2 | Rank | Layer 3 | Rank | Layer 4 | Rank |
|---|---|---|---|---|---|---|---|
| block | 1 | block | 1 | block | 1 | block | 1 |
| blinker | 2 | blinker | 2 | blinker | 2 | blinker | 2 |
| *pond* | *9* | glider | 4 | *pond* | *9* | beehive | 3 |
| beehive | 3 | beehive | 3 | glider | 4 | glider | 4 |
| glider | 4 | pond | 9 | beehive | 3 | boat | 6 |
| ship | 7 | loaf | 5 | boat | 6 | ship | 7 |
| boat | 6 | ship | 7 | ship | 7 | loaf | 5 |
| | | tub | 8 | loaf | 5 | tub | 8 |
| | | boat | 6 | long boat | 10 | pond | 9 |
| | | *pulsar* | *21* | tub | 8 | toad | 11 |
| | | toad | 11 | ship-tie | 12 | pulsar | 21 |
| | | | | | | *loop* | *49* |
| | | | | | | half-bakery | 15 |
| | | | | | | long boat | 10 |
| | | | | | | beacon | 13 |
| | | | | | | mango | 16 |
| | | | | | | ship-tie | 12 |
| | | | | | | barge | 14 |
| | | | | | | integral sign | 25 |
| total types | 7 | total types | 11 | total types | 11 | total types | 19 |
| total freq | 7,799 | total freq | 10,490 | total freq | 10,187 | total freq | 43,865 |
| num seeds | 600 | num seeds | 600 | num seeds | 600 | num seeds | 600 |
| freq / seeds | 13.0 | freq / seeds | 17.5 | freq / seeds | 17.0 | freq / seeds | 73.1 |

Table 4. This table summarizes the correlations between fitness and other attributes of the four layers in Model-S. Fitness is more correlated with diversity than with area or productivity. We evaluate the statistical significance of the correlations using a two-tailed Student t-test for Pearson correlations. All three correlations are statistically significant. The correlations are based on comparing two samples of 48 values each (4 layers × 12 runs).

| Attributes | Correlation | p-value | p-value < 0.05? | Section | Figure |
|---|---|---|---|---|---|
| fitness and area | 0.811 | 2.72E-12 | Yes | 3.1 | Figure 3 |
| fitness and productivity | 0.706 | 2.05E-08 | Yes | 4.1 | Figure 4 |
| fitness and diversity | 0.834 | 1.79E-13 | Yes | 4.2 | Figure 5 |